\documentclass[journal]{IEEEtran}
\usepackage{cite}
\ifCLASSINFOpdf
  \usepackage[pdftex]{graphicx}
  \graphicspath{{../pdf/}{../image/}}
\else
\fi

\usepackage{booktabs}
\begin{document}
\title{Survey of Design Paradigms for Social Robots}

\author{Rita Frieske,
       Xiaoyu Mo,
       Yini Fang,
       Jay Nieles,
       Bertram E. Shi% <-this % stops a space
% \thanks{M. Shell was with the Department
% of Electrical and Computer Engineering, Georgia Institute of Technology, Atlanta,
% GA, 30332 USA e-mail: (see http://www.michaelshell.org/contact.html).}% <-this % stops a space
% \thanks{J. Doe and J. Doe are with Anonymous University.}% <-this % stops a space
}

\maketitle


\begin{abstract}
    The demand for social robots in fields like healthcare, education, and entertainment increases due to their emotional adaptation features. These robots leverage multimodal communication, incorporating speech, facial expressions, and gestures to enhance user engagement and emotional support. The understanding of design paradigms of social robots is obstructed by the complexity of the system and the necessity to tune it to a specific task.  This article provides a structured review of social robot design paradigms, categorizing them into cognitive architectures, role design models, linguistic models, communication flow, activity system models, and integrated design models. By breaking down the articles on social robot design and application based on these paradigms, we highlight the strengths and areas for improvement in current approaches. We further propose our original integrated design model that combines the most important aspects of the design of social robots. Our approach shows the importance of integrating operational, communicational, and emotional dimensions to create more adaptive and empathetic interactions between robots and humans.
\end{abstract}


% \fi
\IEEEpeerreviewmaketitle

\section{Introduction}

Social robots are interactive machines designed to engage with humans in a socially meaningful way. These robots are capable of processing and responding to human emotions, intentions, and actions through various modalities, such as speech, facial expressions, gestures, and other forms of communication. With the increased demand for technological support in fields such as healthcare, education, entertainment, and customer service, a social robot's ability to simulate empathetic social interaction enhances user engagement, emotional support, and task efficiency \cite{Fong2003, Dautenhahn2007}. 

During our work on designing health care robot assistant Grace, we discovered a discrepancy between advancement and a body of literature on traditional chatbots and the coverage of the design and application of social robots. In their survey on dialogue management in HRI \cite{reimann2023survey} point out that the reason behind, it is that most of the dialogue management in HRI is handcrafted based on how the robot is used. We believe that the complexity of the integration of hardware components with the software.
Through our design experience, we discovered that the creation of a plausible design required combining knowledge from various fields stretching far beyond engineering. Subsequently, we decided that there remains a need for a comprehensive analysis of the various design paradigms of social robots, which could be potentially helpful for both researchers and engineers to clarify the scope of their work.

This article contributes to the field by providing a structured review of design paradigms for social robots, categorized into cognitive architecture, role design model, linguistic model, communication flow, activity system model, and integrated design model. By breaking down the articles on social robot design and application based on these paradigms, this study highlights the strengths and areas for improvement in current approaches. Furthermore, our article associates different design paradigms with specific fields of study such as linguistics, psychology, communication theory, neuroscience, and engineering.

Additionally, we propose a novel integrated design model. We discovered that the analyzed designs concentrate either on interactive communicational aspects or engineering aspects when creating the design, while the emotional responses are often added as an additional feature to the model. We believe that since the emotional response is crucial to the social robot operation it has to be considered from the very beginning of the design process. Our integrated design model combines operational, emotional, and communicational dimensions that are crucial for the social robot design.

\begin{table}
\centering

\resizebox{\columnwidth}{!}{%
\begin{tabular}{@{}ll@{}}
\toprule
HRI                                     & NLP                                      \\ \midrule
Multimodal (audio+vision+haptic+others) & Mostly bi-modal (text+image)             \\
Pipeline systems                        & End-to-end deep learning models          \\
Specific domain                         & Open domain or specific domain           \\
Active side: robot or user              & Active side: user                        \\
Wide possibility of output expression   & Limited possibility of output expression \\
Wide context                            & Narrow context (mostly semantic)         \\
Real-time interaction                   & Temporal sequence interaction            \\ \bottomrule
\end{tabular}%
}
\vspace{0.1cm}
\caption{Comparison between HRI and NLP multimodal dialogue systems.}
\label{tab:HRI_NLP}
\end{table}

\section{Features of Dialogue in HRI}

Although communication with robots with natural language as a medium was introduced early on in the robotics field the back and forth communication or more complex interaction is still an underdeveloped task. There is a great discrepancy between the development of NLP dialogue systems and their successful application in robotic agents.

Modern NLP systems are mostly created with computer users in mind, and as such, they work within restrictions of computer communication, namely mostly operating with written text, with no time limits, with only linguistic context and output expression the same as an input. Multimodal dialogue systems in NLP are majority bi-modal, consisting of text and images. Table \ref{tab:HRI_NLP} shows the breakdown of major differences in dialogue in NLP and HRI.

\paragraph{Real-time Interaction}

Time is a more important concept in HRI than in most NLP-powered dialogue systems. Sequential progression is an important concept in natural language, but for robots being immersed in reality the timely and accurate response is crucial for successful operation. This is reflected in different approaches towards state tracking. 

Time is a much more important concept in HRI than in most NLP-powered dialogue systems. Sequential progression is an important concept in natural language, but for robots being immersed in reality a timely and accurate response is crucial for successful operation. This is reflected in different approaches towards state tracking. In the NLP case, the state represents the progression of communication,  focusing primarily on the sequence of conversational turns and the context of the dialogue. The system tracks the history of the conversation to manage and generate appropriate responses, ensuring coherence and relevance in the interaction.

In contrast, for robots, the state is not limited to verbal exchange but also includes real-time sensory inputs and situational awareness. This information may be concentrated on different aspects such as the task at hand, the human interlocutor, or the surrounding environment. For instance, a robot assistant in a healthcare setting needs to process and integrate data from multiple sources—such as patient vitals, environmental conditions, and spoken instructions—to provide timely and contextually appropriate actions.

Moreover, the temporal aspect of HRI necessitates that robots continuously update their state based on immediate inputs and feedback. This includes recognizing and responding to non-verbal cues like gestures, facial expressions, and body language, which can change rapidly and require instantaneous interpretation. This is essential for maintaining the flow of interaction and ensuring the robot's actions are relevant and effective.

This complexity in state tracking for robots also involves managing and prioritizing multiple streams of information simultaneously. For example, a robot may need to balance focusing on a conversational task with monitoring environmental changes that could impact its operation or the safety of humans around it. The integration of multimodal data thus requires sophisticated algorithms capable of real-time processing and decision-making, far beyond the linear progression typically managed in traditional NLP dialogue systems.
%Add Ro_MAn paper 

\paragraph{Wide Communication Context}
One of the distinct features of the application of multimodal pathways in robots is a widening of the communication context. In the case of dialogue systems trained on text, the context is limited to the linguistic cues in a sentence. In the case of ASR systems context consists of acoustic and linguistic cues. HRI widens context that can take additional information from the environment including face tracking, emotion recognition, environment tracking, pose estimation, and pressure cues. This expansion allows robots to gather and interpret a rich array of data, providing a more comprehensive understanding of the interaction and enabling more nuanced responses.

Communication context is not limited to receiving information from the environment. Embodied robots possess more means to mirror and respond to humans with their behaviors and actions. This is especially striking in social robots, due to the need to use gaze, mirroring, and other social cues so that the robot becomes more responsive to users' emotional needs. For instance, a social robot can maintain eye contact, mimic facial expressions, and adjust its posture to align with the emotional state of the human interlocutor, thereby enhancing the sense of connection and empathy in the interaction.

\begin{figure}
    \centering
    \includegraphics[width=0.8\columnwidth]{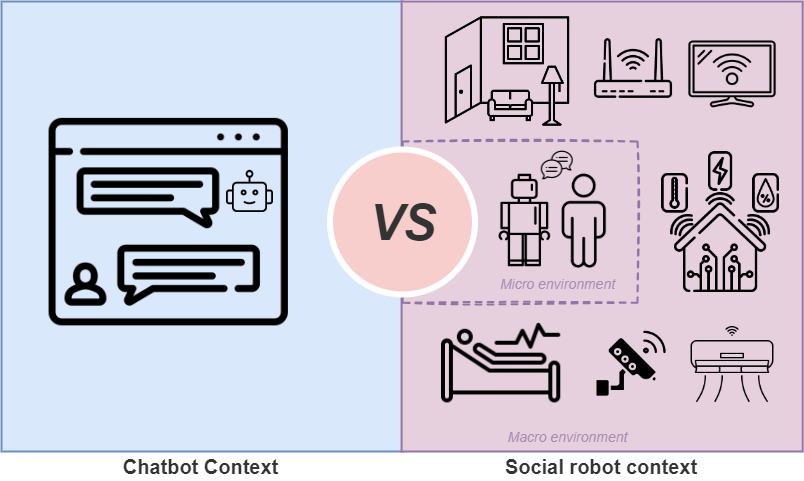}
    \caption{The communication context of text chatbot consists of textual information, whereas for robots that operate in the physical environment, the communication context is grounded in its direct environment and limited only by the number of its sensors. }
    \label{fig:context}
\end{figure}

\paragraph{Symmetrical Communication}

The aspect very specific to social-robot interaction is their active role in the communication process. Personal assistants such as Apple Siri or Amazon Alexa are passively waiting for the user's command, which is clear since the user needs to use a "wake up phrase" to start the interaction. In the case of a social robot, both humans and robots can be agents, since the task of some social robots is engaging humans to interact (such as in the case of humanoid elderly companions). This ability to initiate interaction allows social robots to participate in a more symmetrical communication pattern, where the flow of dialogue is more dynamic and can be influenced by both parties.

Social robots are designed to exhibit social behaviors and cues that facilitate a bidirectional flow of communication. Robots can initiate interactions by moving closer to a person, gesturing to attract attention, or even using tactile feedback, such as a gentle touch, to provide comfort or reassurance. These behaviors are not only crucial for making the interaction feel more human-like but also for ensuring that the robot can effectively engage and maintain the user's interest. In contrast, virtual chatbots typically lack these physical attributes and rely solely on textual or vocal exchanges, which can limit their ability to create a symmetrical experience and put them in a more passive role.

\paragraph{Multimodality}
The concept of multimodality is natural to HRI due to different sensory inputs. Recently this idea was transplanted into deep learning frameworks including multimodal dialogue in NLP. The variety of fields where the concept is used caused it to become blurry. For this survey, we assume the definition of the modality as pathways of sensory experience for the perceiving entity.
\begin{figure}
    \centering
    \includegraphics[width=0.8\linewidth]{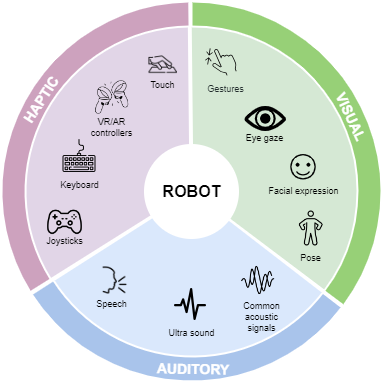}
    \caption{Diagram of auditory, haptic, and visual modalities with examples of processed information.}
    \label{fig:multimodal}
\end{figure}

In human-robot interaction, different modalities can be used to convey different types of information. For example, speech can be used to communicate complex ideas and instructions, while gesture can be used to indicate simple commands or express emotions. Touch can be used to provide feedback or convey a sense of physical presence, while facial expressions can be used to communicate emotions and attitudes.
In their survey \cite{Su2023} compile the commonly used modalities in Human Robot interactions divided by input and output. They order main modalities understood as information pathways into visual, auditory, haptic, kinesthetic, and proprioceptive. We describe the three former ones since they have a direct impact on the multimodal dialogue systems. Kinesthetic ability is the ability of a robot to sense and decode motion and movement and proprioception is the ability of an organism to sense the position and orientation of its body parts. These two modalities are currently more relevant to industrial robots and exoskeletons rather than dialogue systems thus we do not analyze them in this survey. 

There is a growing body of work on multi-modal dialogue systems, however, the majority of the cases concern visual dialogue such as in~\cite{sun-etal-2022-multimodal, he2020}, which is the task that takes in images and text as inputs and produces text or both text and images in the response turn, the most famous example of such system nowadays is GPT4~\cite{gpt4}. Several works mention dialogue incorporating modalities different from static images and text such as from \cite{frieske2024erit}. After text and image, the next popular modality is audio and speech. The works on video and dialogue resulted in a visual dialogue dataset that incorporates gaze tracking to dialogue systems \cite{kamezawa-etal-2020-visually}, the feature crucial for social robot interaction.

\section{Design}

The design of multimodal dialogue systems in robots largely adjusts itself to the task of the robot, and so the majority of the implementation falls into the NLP category of task-oriented dialogue. As of now, almost all robotic systems use a kind of framework that allows for the integration of multiple sensors and adjustment to the performed task rather than training it end-to-end. 
Since the design of a social robot is an interdisciplinary undertaking we propose a framework based on analysis of multiple design approaches that allows us to treat the dialogue design process as a path through different paradigms. We propose looking at the dialogue design not only as an engineering task, but also include communication, psychological, and linguistic perspectives. Our approach is based on the survey of the literature as well as our experience in the design process of the social robot Grace.

% \begin{table*}[]
% \centering
% \resizebox{\textwidth}{!}{%
% \begin{tabular}{@{}llll@{}}
% \toprule
% Design Paradigm       & Examples               & Field & Articles \\ \midrule
% Cognitive Robotics & Standard Model, Soar, ACT-R, DUAL, CLARION & Neuroscience, Psychology, Computing, Robotics &  \cite{ laird2017standard, Laird2012, Ritter2018, Jokinen2018HumanRobotDF}      \\

% Communication Flow    & Constructive Dialogue Model  & Communication    &  \cite{Jokinen2018,Jokinen2018HumanRobotDF}           \\
% Linguistic Model      & Communicative Acts, Acyclic Graphs   & Linguistics &   \cite{Rodicio2020Modelling , Berzuk2022}        \\
% Role Design Model  &                   & Communication    &    \cite{Mehlmann2016, Frijns2021}      \\
% Activity System Model &                 &  Psychology      &    \cite{Tewari2022, Lala2021FindingsFH}      \\ 
%  Integrated Design Model &                 &  Engineering, Psychology, Linguistics      &    \cite{Tewari2022, Lala2021FindingsFH}      \\ 

% \bottomrule
% \end{tabular}%
% }
% \end{table*}

\subsection{Cognitive Architectures} is a group of models that try to create a robot cognition framework based on neuroscience. Some of the examples of cognition models are the standard model of the mind (or standard model) proposed first by Laird et al. in 2013 \cite{laird2017standard}, ACT-R proposed by \cite{Ritter2018}, SOAR \cite{Laird2012}, CLARION or DUAL. Cognitive architectures provide a structured approach to developing intelligent systems by mimicking human cognitive processes. They enable robots to perceive, reason, and act in complex environments, supporting advanced dialogue capabilities.

Standard Model of the Mind: This model aims to capture general cognitive functions such as perception, memory, learning, and decision-making, creating a comprehensive framework for understanding and simulating human cognition.
ACT-R (Adaptive Control of Thought-Rational): ACT-R is a cognitive architecture that simulates human thought processes in a structured and hierarchical manner, focusing on the interaction between declarative and procedural knowledge.
SOAR: SOAR is designed to model general intelligence through goal-directed behavior, using a problem-space computational model to simulate human reasoning and learning.
CLARION (Connectionist Learning with Adaptive Rule Induction ON-line): CLARION integrates symbolic and subsymbolic processes to model human cognition, emphasizing the role of implicit and explicit knowledge in learning and decision-making.
DUAL: DUAL is a cognitive architecture that combines rule-based and connectionist approaches, aiming to capture the parallel and interactive nature of cognitive processes.

During face-to-face interactions with robots personalised approaches and remembering user data and new words account for fluent and natural interaction.
\cite{Amiri2019} explore possibilities of statistical updates of knowledge databases in dialogue systems that allow integrating user information and new terms, exploring the

\subsection{Role Design Model} is the design paradigm that designs actions of both humans and robots in the interaction. Conversely, to certain systems that treat humans as a 'black box' that can create every possible scenario, this model looks into the task that the robot has to perform and tries to create a finite number of interaction paths for both humans and robots. It takes inspiration from the communication theory and the idea that interaction with the robot can be symmetrical and asymmetrical \cite{Frijns2021}. However the idea comes naturally from the need to design the script of the interaction itself before getting into the details of the architecture, it accepts the user as an integral part of the dialogue design rather than treating him only as a creator of an input.

\cite{Mehlmann2016} look at the dialogue system as a series of actions assigned to humans and to robots rather than only potential robotic responses. By looking at the dialogue act from both human and robot perspective they assign certain roles to both parties in the same way drawing the limits of the interaction.

In their 2022 survey \cite{Berzuk2022} analyzed 75 articles on robotic dialogue systems and categorized dialogue design along 3 axes: interlocutor composition, dialogue structure, and discourse genres. Interlocutor composition means the number of robots and humans interacting. Dialogue structure informs whether dialogue is scripted and linear, can be more flexible and branch out, or finally depends on user input and be open domain and unstructured. Discourse genres finally are their linguistic approach that divides dialogue intentions into commanding, questioning, informing, and entertaining.

\subsection{Linguistic Model}

This approach views speech as an action rather than just conveying information. It accommodates the initiative for communication from both the user and the robot, categorizing actions into giving and receiving information. The Linguistic Model focuses on the pragmatic aspects of language use, such as speech acts, conversational implicatures, and context-dependent meaning.

Speech Acts: Speech acts theory categorizes utterances based on their function in communication, such as asserting, questioning, requesting, or commanding.
Conversational Implicatures: This concept refers to the implied meanings and intentions behind spoken language, which are often inferred from context and shared knowledge.
Context-Dependent Meaning: The meaning of an utterance can vary based on the situational context, requiring the system to interpret language dynamically and adaptively.
\cite{Rodicio2020Modelling} introduce so-called communicative acts (CA) and complex communicative acts (CCA) as basic interaction units modeling dialogue between the robot and the user. Their approach concentrates on linguistic understanding of speech as an action rather than just a form of conveying information. CAs accommodate the fact that initiative for communication can come both from the user and from the robot, which differentiates social robots from assistants such as Google or Alexa that do not take initiative. Furthermore, CAs are divided into those giving and receiving information.

\begin{table*}

\resizebox{\textwidth}{!}{%
\begin{tabular}{@{}llll@{}}
\toprule
Design Paradigm &
  Examples &
  Approach &
  Articles \\ \midrule
Cognitive Architectures &
  \begin{tabular}[c]{@{}l@{}}Standard Model, Soar, \\ ACT-R, DUAL, CLARION\end{tabular} &
  \begin{tabular}[c]{@{}l@{}}Neuroscience, \\ Psychology, Robotics\end{tabular} &
  \begin{tabular}[c]{@{}l@{}}\cite{laird2017standard}, \cite{Ritter2018},\\  \cite{Laird2012}\end{tabular} \\
Role Design Model &
   &
  Communication Theory & 
  \begin{tabular}[c]{@{}l@{}} \cite{Frijns2021}, \cite{Berzuk2022} ,\\ \cite{Mehlmann2016}\end{tabular} \\ 
Linguistic Model &
  \begin{tabular}[c]{@{}l@{}}Communicative Acts, \\ Acyclic Graphs\end{tabular} &
  Linguistics &
  \cite{Rodicio2020Modelling} \\
Communication Flow &
  \begin{tabular}[c]{@{}l@{}}Constructive Dialogue Model, \\ Dialogue Manager\end{tabular} &
  Computing, Communication Theory &
  \begin{tabular}[c]{@{}l@{}}\cite{Jokinen2018}, \cite{Jokinen2018HumanRobotDF}, \\ \cite{reimann2023survey}, \cite{Abbasi},\\  \cite{Lala2021FindingsFH}, \cite{ishii-etal-2021-erica}, \\ \cite{Amiri2019}, \cite{Milhorat2018ACD}\end{tabular} \\
Activity System Model &
   &
  Psychology &
  \begin{tabular}[c]{@{}l@{}}\cite{leont1978activity}, \cite{Tewari2022},\\ \cite{Abbasi} , \cite{Amiri2019}\end{tabular}\\
Integrated Design Model &
   &
  \begin{tabular}[c]{@{}l@{}}Psychology, Communication Theory, \\ Robotics\end{tabular} &
  \begin{tabular}[c]{@{}l@{}}\cite{shi2019review}, \cite{chen2013review}, \\ \cite{wang2022reducing}, \cite{wang2023designing}, \\ \cite{baghaei2021virtual}, \cite{Sakakibara2017},\\ \cite{Bickmore2005}, \cite{Shibata2011}, \\ \cite{Wright2023}, \cite{jean2008design},\\ \cite{liu2016multimodal}, \cite{johnson2016exploring}, \\ \cite{foster2016mummer}, \cite{liang2023robot},\\ \cite{majgaard2015multimodal}, \cite{van2020teachers}\end{tabular} \\ \bottomrule
\end{tabular}%
}
\vspace{0.1cm}
\caption{Breakdown of the articles on social robot design and application based on the categorized design paradigms. This table categorizes cited authors from the article according to their respective design paradigms, highlighting key contributions to the development and application of multimodal dialogue systems.}
\label{tab:design}
\end{table*}

\subsection{Communication Flow}

Communication flow provides a framework for managing the exchange of information in a dialogue, ensuring that interactions are coherent, contextually relevant, and goal-oriented.

Awareness and Perception: Both the human and the robot need to be aware of each other's communication attempts and perceive multimodal signals accurately.
Understanding and Interpretation: The system must be capable of interpreting the meaning of signals and utterances based on context, previous interactions, and multimodal cues.
Response and Reaction: The robot's responses should be timely, relevant, and appropriate to the ongoing dialogue, maintaining the flow and coherence of the interaction.

\cite{Jokinen2018, Jokinen2018HumanRobotDF} look at the dialogue from the perspective of roles and capabilities of robot and user, i.e. what human and robot should be able to do within the interaction. Building on this notion they use Constructive Dialogue Model (CDM). In this model, conversation progresses cyclically with encouragement to continue conversations. Both parties need to be in contact and aware of the partner's attempt to communicate. They need to perceive all the multimodal signals, understand their meaning, and react appropriately. Therefore modules of such a dialogue system are divided into contact, perception, understanding, and reaction modules. This system was specifically designed for humanoid Kristina, based on work with another humanoid, ERICA, \cite{Milhorat2018ACD} propose a question-answering system with a fallback to generate more naturally sounding dialogues.

In their survey \cite{reimann2023survey} look at the human-robot dialogue specifically from the perspective of dialogue management. Dialogue management controls turn-taking between users and robots, that is controls communication exchange between them.

\subsection{Activity System Model}

Derived from activity theory, this model focuses on the purposeful activity defined by its object. It emphasizes the use of mediating tools to affect and transform the activity's object, providing a deeper understanding of human-robot interaction. The Activity System Model highlights the dynamic and situated nature of dialogues, considering the broader context of activities and social interactions.

Activity Theory: Activity theory, first proposed by \cite{leont1978activity}, provides a framework for analyzing human actions as purposeful and goal-directed, influenced by social, cultural, and material factors. The goal of the activity is to fulfill an individual’s needs. 
Mediating Tools: These tools, such as language, gestures, and artifacts, mediate the interaction between the human and the robot, facilitating communication and task completion.
Object of Activity: The object of the activity defines the purpose and direction of the interaction, shaping the roles, actions, and outcomes of the dialogue.
 This interplay between the three core components of human activity: subject, object, and tool is often visualized in the form of a triangle. Through their analysis of older and middle-aged adults' interaction with robots \cite{Tewari2022} discovered four themes influencing HRI: expecting, understanding, relating, and interacting. The themes span complex human activity at different complementary levels and provide a further developed understanding of breakdown situations in human-robot interaction (HRI). In their experiments, older and middle-aged adults emphasized emphatic behavior and adherence to social norms, while younger adults focused on functional aspects such as gaze, response time, and length of utterances.
 
\cite{Abbasi} integrates pointing information with utterance and passes it into the mediation module that defines actions that the robot needs to take, which falls into the category of activity theory.

In their work with humanoid ERICA \cite{Lala2021FindingsFH} analyzes dialogue system performance for ERICA in several social roles to establish the extent to which the robot could be used formally.

% \begin{figure*}[]
%     \centering
%     \includegraphics[width=0.8\textwidth]{image/designParadigm0206.png}
%     \caption{Diagrams of different design paradigms}
%     \label{fig:enter-label}
%     % changing comments: 
%     % cognitive robotics(standards system model)
%     % communication flow, human-robot (symmetric communication model)
%     % liguistic model, activity system model
%     % acyclic graph, dialogue system(architectural model). 
%     % direct use: dialogue system, acyclic graph, 
%     % will find later: activity system, 
%     % figure out: linguistic model, human-robot, communication flow, cognitive robotics. 

%     % context image: delete human icon(left), right: user and robot is talking, 
% \end{figure*}

\subsection{Our Novel Paradigm: Integrated Design Model}
Our novel paradigm combines the operational dimension, emotional dimension, and communicational dimension of design, based on knowledge distilled from the analyzed articles. This integrated approach ensures a holistic design of multimodal dialogue systems for social robots, balancing functionality with user experience.

\paragraph{Operational Dimension} includes the technical and functional components required for the robot to perform tasks efficiently. It encompasses sensor integration, real-time processing, and task-specific adaptations to ensure the system's operational efficiency and reliability.
The most important question asked while designing a robot is what is the task it is going to perform. Hence, task adaptation customizes the system's behavior and responses based on the specific tasks and contexts in which the robot operates, enhancing its effectiveness and relevance.
The second important element of the operational dimension is module integration. It is responsible for combining inputs from various sensors such as cameras, microphones, and touch sensors to provide a comprehensive understanding of the environment and user interactions.
The real-time processing aspect of operation ensures that the system can process multimodal inputs quickly and accurately to respond in real-time, which is essential for maintaining natural and fluid interactions.

\paragraph{Emotional Dimension}  focuses on the robot's ability to recognize, interpret, and respond to human emotions, making interactions more engaging and empathetic.
Emotion Recognition: Utilizing facial expressions, voice tone, and body language to detect and interpret the user's emotional state.
One of the most important design aspects in this dimension is designing the robot's responses to acknowledge and appropriately address the user's emotions, fostering a sense of understanding and connection.
By modifying the robot's behavior and interaction style based on the user's emotional state, we can ensure a supportive and positive interaction experience. Conceptualizing mirroring strategies is essential for a social robot to showcase its empathetic capabilities.

\paragraph{Communicational Dimension} includes the design of the dialogue flow, linguistic processing, and multimodal interaction management to ensure effective and natural communication.
Design of the dialog flow means structuring the conversation to be coherent, contextually relevant, and goal-oriented, maintaining the user's engagement and satisfaction. Turn-taking and barging in are elements of communication that need to be conceptualized on this level of the design.
Handling natural language understanding and generation to interpret user input accurately and generate appropriate responses depends on linguistic processing, this consists of elements such as semantic parsing.
Finally, by coordinating inputs and outputs across different modalities (e.g., speech, gestures, touch), we create a seamless and intuitive interaction experience. Hence multimodal fusion is an element that appears in this design level.
        
% \paragraph{Examples of Integrated Design Model Paradigms}

% Emotionally Aware Task Assistant
%     Operational Aspects: The system integrates visual and auditory sensors to recognize tasks and user commands in real-time, adapting its actions based on the specific tasks it needs to perform.
%     Emotional Aspects: It uses facial recognition and voice analysis to detect user emotions, providing empathetic responses and adjusting its behavior to support the user emotionally during task execution.
%     Communicational Aspects: The dialogue flow is designed to be supportive and encouraging, using natural language processing to understand and respond to user inputs effectively across multiple modalities.

% Adaptive Social Companion
%     Operational Aspects: The robot combines environmental sensors and contextual data to understand the user's environment and activities, adjusting its actions and interactions accordingly.
%     Emotional Aspects: It continuously monitors the user's emotional state through multimodal cues, offering companionship and emotional support tailored to the user's needs.
%     Communicational Aspects: The system manages multimodal interactions, coordinating speech, gestures, and touch inputs to create a natural and engaging conversational experience, promoting user well-being and social interaction.

\section{Evaluation of Social Robot Designs}

Just as with robotic design, a comprehensive evaluation of robotic dialogue systems is a complex task due to the software and hardware involved, as well as depending on the task that the robot is dealing with. There exist several evaluation frameworks designed for robots but they refer to the interaction of operators with factory robotic interfaces \cite{Scholtz2003,Apraiz2023}. Due to the system complexity, there is a prevalence of qualitative evaluation standards in robotics, where the human subjects evaluate what is their impression of the particular robot \cite{Sirithung}. The quantitative approach mostly assesses metrics from particular submodules.

\subsection{Quantitative Evaluation}
According to \cite{Deriu2021} automatic evaluation of the dialogue system should be repeatable, correlated to human judgments, should differentiate between different dialogue systems, and should be explainable. Some evaluation metrics aim to evaluate task-oriented dialogue systems, hence the main objective is evaluating the success rate of the task i.e., successful non-interrupted interaction. Such metrics include the number of turns needed for the successful interaction, including information about the backchanneling turns that result from barging in. Finally, it combines the error rates from modules processing different modalities, such as ASR or CV to evaluate their impact on the overall system. 

\cite{Abbasi} in their work concerning task-oriented dialogue with pointing and object recognition combine qualitative and quantitative evaluation. In their work they evaluated qualitatively system as an entity using the rate of successful interactions as well as particular modules, e.g. they used word error rate (WER) for speech recognition as well as their measure of serious speech recognition errors (SSRE) that was defined as speech errors that are alter the meaning of the sentence so that task cannot be completed. Pointing errors were shown by the percentage of not-recognized gestures or the ones that were not tagged with the correct location. The performance of the multimodal fusion module was evaluated by comparing generated dialogue acts tags with manually labeled dialogue acts of all utterances.

For the dialogue module alone the most standard metrics are usually F1, accuracy, or precision. \cite{ni2023} differentiate the dialogue evaluation between open domain and task-oriented dialogue. The former type is mostly evaluated with BLEU, Rouge, and METEOR. The latter can be evaluated by the Task Completion Rate or Task Completion Cost. Task-oriented dialogue is the dialogue type mostly implemented in social robots, and the Task Completion Rate can be easily applicable to most of the use cases.  \cite{yeh-etal-2021-comprehensive} test multiple automatic dialogue metrics, however apart from BLEU, Rouge, and METEOR, many of them are model dependent i.e. they are designed to evaluate BERT-based models.
There is a growing body of work on multi-modal dialogue systems, however, the majority of the cases concern visual dialogue such as in~\cite{sun-etal-2022-multimodal, he2020}, who use F1, BLEU, NIST, Rouge, perplexity, and image specific Frechet Inception Distance (FID) and Inception Score (IS). 

The combination of many subsystems creates a danger of error propagation in social robots, that is the possibility that errors from one component will influence the performance of others. Furthermore, with the common deployment of deep learning models, errors such as hallucinations might become a real threat \cite{10.1145/3571730}, even more if they are not detectable by the common metrics such as hallucination is ASR~\cite{frieske2024hallucinationsneuralautomaticspeech}.

\subsection{Qualitative Evaluation}

For qualitative robot evaluation, the most widely used questionnaire is based on the Godspeed questionnaire \cite{Bartneck}.  Often unrelated questionnaires are used to evaluate aspects not mentioned in Godspeed.  For instance, one might add aspects of trust evaluation \cite{Zhu22}, since this is the element that is not used in the Godspeed questionnaire. 

Human evaluation encompasses the subjective assessment of dialogue systems from the perspective of users. Irrespective of the specific approach employed, all human evaluations involve soliciting input from external annotators who respond to inquiries regarding dialogues.

Previous studies have exhibited considerable variation in the dimensions used to evaluate dialogue systems through human evaluations. For instance, certain studies \cite{tian2019learning, zhou2019unsupervised} focus on assessing the overall quality of dialogue system responses. \cite{moghe2018towards} examines the "Humanness" of responses, capturing the extent to which they appear human-like. Emotion \cite{li2018syntactically} and Empathy \cite{lin2019moel} dimensions encompass the evaluation of both the understanding and generation of emotional responses. Furthermore, \cite{zhang2018personalizing} directs attention to "engagingness", evaluating whether the response includes interesting content. Proactivity, used in \cite{wu2019proactive}, pertains to whether the response introduces new topics while maintaining coherence. \cite{liu2018knowledge, wang2020improving} examined accuracy, which measures the correctness of responses based on real-world knowledge, and, knowledge relevance \cite{liu2018knowledge, wang2020improving}, which investigates whether the knowledge conveyed in the response aligns appropriately with the context.

\section{Integrated Design for Social Robot Applications}
% Applications of multimodal dialogue systems in human-robot interaction:

The application of social robots spans fields that benefit from empathetic mirroring in technology. In this section, we list current and possible deployments of functioning social robots and analyze them through the perspective of our integrated design model to showcase their most important elements: functionality, emotional responsiveness, and communicational agility.

\subsection{Healthcare}
Social robots can be used in the field of healthcare, for example, to provide emotional support for patients. A robot that can respond to a patient's emotional cues, such as facial expressions and gestures, can provide comfort and companionship for the patient, which can be especially important for patients who are isolated or lonely. Wright's anthropological analysis describes how robots influence elderly care in Japan, noting that the operation of these robots can be time-consuming for nurses \cite{Wright2023}.

In the contemporary healthcare landscape, technology plays an increasingly pivotal role, exemplified by innovations such as exoskeletons in rehabilitation \cite{shi2019review, chen2013review} and the application of Virtual Reality (VR) in psychological care \cite{wang2022reducing, wang2023designing, baghaei2021virtual}. Furthermore, social robots, which integrate various forms of communication, are gaining traction in healthcare.

\paragraph{Operational Dimension} Social robots in healthcare integrate technical and functional components necessary for task efficiency, such as sensor integration, real-time processing, and task-specific adaptations. These systems enhance user engagement by interpreting cues from various user groups, including the elderly and individuals with dementia, and responding through multiple modalities. For instance, exoskeletons and rehabilitation tools are rooted in the physical embodiment of the robot, enhancing the operational capacity in therapeutic settings \cite{shi2019review, chen2013review}. Sakakibara et al. developed a humanoid virtual caregiver capable of interacting with dementia patients using voice, incorporating life history and linked open data to facilitate person-centered conversations \cite{Sakakibara2017}.

\paragraph{Emotional Dimension} Social robots excel in recognizing, interpreting, and responding to human emotions, providing emotional support, cognitive training, and soothing companionship to users. Bickmore et al. observe that with the decline of mental capabilities and sensory functions in the elderly, one skill that people often retain is multimodal face-to-face communication \cite{Bickmore2005}. Shibata et al. introduced a unique robot with a seal-like appearance, designed to interact with and provide companionship to the elderly, featuring soft artificial fur that evokes a sense of warmth and comfort \cite{Shibata2011}. 

\paragraph{Communicational Dimension} These systems design dialogue flow, linguistic processing, and multimodal interaction management for effective and natural communication. In their work  \cite{ishii-etal-2021-erica} explored humanoid ERICA as a potential well-being assistant with emotion recognition capacity for maintaining communication with patients during COVID-19.

\subsection{Entertainment}
% One of the most common applications is in the field of entertainment, such as interactive games and toys. For example, a robot that can respond to a user's speech and gesture inputs can provide a more engaging and enjoyable gaming experience.

In the domain of entertainment applications, we primarily observe two distinct scenarios.

\paragraph{Operational Dimension} Technical and functional components include the integration of sensors and real-time processing for interactive play with robots. For instance, Jean et al. investigated a robot designed to exhibit behaviors such as singing, crying, laughing, and looking around, focusing on user interactions for general entertainment purposes \cite{jean2008design}. Liu's study on playing finger guessing with a robot also highlights the operational capabilities \cite{liu2016multimodal}.

\paragraph{Emotional Dimension} Social robots in entertainment recognize and respond to human emotions to enhance user engagement. Johnson et al. demonstrate that entertainment experiences are enriched when robots provide expected multimodal feedback during games, adding a new dimension to the entertainment value \cite{johnson2016exploring}.

\paragraph{Communicational Dimension} Effective and natural communication is achieved through the design of dialogue flow, linguistic processing, and multimodal interaction management. This is exemplified by the project MuMMER (MultiModal Mall Entertainment Robot), led by Foster et al., where a humanoid robot engages with customers in a shopping mall through a combination of speech-based interaction, non-verbal communication, and human-aware navigation \cite{foster2016mummer}.

\subsection{Education}
Another application of multimodal dialogue systems is in the field of education.

\paragraph{Operational Dimension} Robots can be used as teaching assistants, integrating sensor inputs, real-time processing, and task-specific adaptations to help students learn new concepts and skills. Liang et al. demonstrated that a robot digital storytelling approach based on multimodal dialogue systems can facilitate students' speaking and storytelling abilities better than traditional one-way videos \cite{liang2023robot}.

\paragraph{Emotional Dimension} These systems recognize and interpret student emotions to make learning interactions engaging and empathetic. The NAO robot, used in a workshop led by Majgaard Gunver, provided an engaging learning experience for students aged 11 to 16, teaching various subjects such as programming, language, ethics, technology, and mathematics. Students enjoyed the interactive experience, which enhanced their knowledge beyond the core curriculum \cite{majgaard2015multimodal}.

\paragraph{Communicational Dimension} The design of dialogue flow, linguistic processing, and multimodal interaction management ensures effective and natural communication. Gijs van Ewijk and colleagues examined the ethical considerations of social robots in educational settings, discussing privacy, security, psychological well-being, and user-friendliness through targeted interviews with teachers. This research provides an in-depth analysis of both the challenges and opportunities in the development and design of educational robots \cite{van2020teachers}.

\section{Challenges and Conclusion}

The primary challenges in multimodal dialogue systems revolve around effectively managing complex interactions across diverse domains, including semantic parsing, turn-taking, and facial expression recognition. One key difficulty is switching between task-oriented and open-domain interactions, which requires a careful balance to avoid the rigidity of task-specific systems and the potential biases of open-domain models. Semantic parsing presents its challenges due to the variability and ambiguity of natural language inputs, which can be compounded by multimodal inputs that may conflict \cite{Zhang2020}. Turn-taking involves accurately determining when and how to shift communication between humans and robots, especially when multiple modalities are involved \cite{liu2016multimodal}. Facial expression recognition faces hurdles in achieving real-time accuracy in dynamic, real-world settings and integrating subtle micro-expressions that reveal genuine emotional states \cite{fang2023rmes}.

In addressing these challenges, incorporating various design approaches can enhance the effectiveness and robustness of social robots. Balancing between task-oriented and open-domain interactions involves integrating both rigid and flexible paradigms to optimize performance and safety. Design approaches such as hybrid models that blend neural and rule-based methods can address this challenge by combining the structured reliability of task-oriented systems with the adaptability of open-domain models. Data augmentation techniques, as suggested by \cite{Zhang2020}, can further support this balance by generating diverse training scenarios that help systems handle multiple correct responses and minimize biases. By leveraging these design approaches, systems can navigate the complexities of human interactions while maintaining accuracy and relevance.

Semantic parsing, turn-taking, and facial expression recognition all benefit from integrating specific design paradigms. For semantic parsing, employing linguistic models and cognitive architectures can enhance the system's ability to interpret and integrate multimodal inputs accurately. Techniques such as dependency parsing and semantic role labeling, combined with multimodal data fusion, improve the system’s understanding of complex user inputs. Turn-taking challenges can be addressed by analyzing communication flow models that ensure smooth transitions and natural interactions. In terms of facial expression recognition, taking an integrated design model as a starting point allows for better planning of real-time processing and interpretation of expressions, leading to more empathetic interactions. These design-focused strategies help tackle the inherent difficulties in multimodal dialogue systems, creating more effective and engaging human-robot interactions.

In conclusion, the design of social robots must integrate a variety of paradigms to effectively address the complex challenges of multimodal communication. By focusing on operational, emotional, and communicational dimensions, designers can enhance the functionality and user experience of robots. The operational design ensures that robots perform tasks efficiently through sophisticated sensor integration and real-time processing. Emotional design contributes to more engaging and empathetic interactions by enabling robots to recognize and respond to human emotions. The communicational design facilitates natural and effective dialogue by managing the flow of conversation, linguistic processing, and multimodal interactions. Addressing these dimensions holistically allows for the creation of social robots that are not only task-oriented but also capable of delivering nuanced and intuitive interactions. As research and technology continue to evolve, incorporating these diverse design approaches will be essential in advancing social robots that connect more deeply with users, making human-robot interactions both more functional and more human-centered.

\newpage

% 



\bibliographystyle{apalike}
\bibliography{bib/dialog, bib/fusion, bib/social_robot_applications, bib/general_dialogue, bib/multimodal, bib/IEEEexample}
\end{document}